\documentclass{article}

\PassOptionsToPackage{numbers, compress}{natbib}
     \usepackage[preprint]{neurips_2019}

\usepackage[utf8]{inputenc} % allow utf-8 input
\usepackage[T1]{fontenc}    % use 8-bit T1 fonts
\usepackage{hyperref}       % hyperlinks
\usepackage{url}            % simple URL typesetting
\usepackage{booktabs, multirow}       % professional-quality tables
\usepackage{amsfonts}       % blackboard math symbols
\usepackage{nicefrac}       % compact symbols for 1/2, etc.
\usepackage{microtype}      % microtypography

\usepackage{graphicx}
\usepackage{amsmath}

\title{Recurrent Connectivity Aids Recognition of Partly Occluded Objects}

\author{%
  Markus Roland~Ernst\\
  Frankfurt Institute for Advanced Studies\\
  and Goethe-Universit\"at Frankfurt\\
  60438 Frankfurt am Main \\
  \texttt{mernst@fias.uni-frankfurt.de} \\
  \And
  Jochen Triesch \\
  Frankfurt Institute for Advanced Studies\\
  and Goethe-Universit\"at Frankfurt\\
  60438 Frankfurt am Main \\
  \texttt{triesch@fias.uni-frankfurt.de} \\
  \AND
  Thomas Burwick \\
  Frankfurt Institute for Advanced Studies\\
  and Goethe-Universit\"at Frankfurt\\
  60438 Frankfurt am Main \\
  \texttt{burwick@fias.uni-frankfurt.de} \\
}

\begin{document}

\maketitle

\begin{abstract}
Feedforward convolutional neural networks are the prevalent model of core object recognition. For challenging conditions, such as occlusion, neuroscientists believe that the recurrent connectivity in the visual cortex aids object recognition. In this work we investigate if and how artificial neural networks can also benefit from recurrent connectivity. For this we systematically compare architectures comprised of bottom-up (B), lateral (L) and top-down (T) connections. To evaluate performance, we introduce two novel stereoscopic occluded object datasets, which bridge the gap from classifying digits to recognizing 3D objects. The task consists of recognizing one target object occluded by multiple occluder objects. We find that recurrent models perform significantly better than their feedforward counterparts, which were matched in parametric complexity. We show that for challenging stimuli, the recurrent feedback is able to correctly revise the initial feedforward guess of the network. %and we visualize how feedback connections influence the perceived input stimulus by back-propagating the mismatch in hidden activations over time to the input layer. 
Overall, our results suggest that both artificial and biological neural networks can exploit recurrence for improved object recognition.
%  The abstract paragraph should be indented \nicefrac{1}{2}~inch (3~picas) on
%  both the left- and right-hand margins. Use 10~point type, with a vertical
%  spacing (leading) of 11~points.  The word \textbf{Abstract} must be centered,
%  bold, and in point size 12. Two line spaces precede the abstract. The abstract
%  must be limited to one paragraph.
\end{abstract}

\section{Introduction}
The primate visual system is able to correctly identify an object within 200~ms of its presentation \citep{thorpe1996speed, potter1976shortterm}. Given this rapidness, object recognition has been widely assumed to be a purely feedforward process \citep{dicarlo2012visualobject}. This idea has been corroborated by the recent success of feedforward convolutional neural networks in the realm of object recognition \citep{krizhevsky2012imagenet, lecun2015deeplearning}. However, both anatomical and physiological evidence point to the importance of recurrent connectivity within this process. The densities of feedforward and recurrent connections in the ventral visual pathway are comparable in magnitude \citep{felleman1991distributed, sporns2004small}, and electrophysiological experiments have shown that processing of object information evolves in time, beyond what would normally be attributed to a feedforward process \citep{cichy2014resolving, brincat2006dynamic, rajaei2019beyond}. In particular, \citep{johnson2005recognition, tang2014spatiotemporal} observed that recognition of degraded or occluded objects produces delayed behavioural and neural responses, which are believed to be a result of competitive processing within lateral recurrent connections \citep{adesnik2010lateral}. Late phase IT response patterns have been shown to be reliably predicted only by shallow recurrent networks and very deep feedforward networks \citep{kar2019evidence}.

For partially occluded images, \citep{smith2010nonstimulated, tang2018completion, tang2014spatiotemporal} suggest that recurrent top-down connections are able to reconstruct missing information. Whether object recognition in artificial neural networks can benefit from recurrent connections is less clear, however. Early investigations of this question used highly restricted datasets, where artificial inputs were partly faded out or masked \citep{spoerer2017recurrent, oreilly2013recurrent}. Under natural conditions, however, occlusion is highly dependent on viewing angle and primates perceive it stereoscopically with two eyes. Thus we here developed two novel occluded image datasets that capture the full range of disparity and perspective cues for both natural (handwritten digits) and computer rendered (full 3-D objects) stimuli.

%This claim is further supported by research showing improved classification performance on occluded stimuli using recurrent convolutional neural networks \citep{spoerer2017recurrent, liang2015recurrent}. However, the stimuli used in previous studies hardly resemble occlusion as it would occur naturally \citep{oreilly2013recurrent, smith2010nonstimulated, tang2014spatiotemporal, wyatte2012limits, spoerer2017recurrent}. Occlusion is highly dependent on viewing angle and primates perceive it stereoscopically with two eyes. Rather than fading out part of the input image or simply masking the input with another one, we set out to investigate the effects of occlusion in a more natural setting. Thus, we developed two novel occluded image datasets that capture the full range of disparity and perspective cues.

We test and compare a range of recurrent convolutional neural networks with different connection properties. Assuming the naming scheme of \citep{spoerer2017recurrent}, we distinguish bottom-up (B), top-down (T) and lateral (L) connections. Bottom-up and top-down correspond to information processing from lower and higher regions, whereas lateral connections process information within the same region of the ventral visual hierarchy. To test whether recurrent networks outperform their feedforward counterparts in a natural occlusion scenario, the network architectures were tasked with classifying objects under different levels of occlusion. Our results show significant performance gains for recurrent architectures. Finally, we explore how recurrent connections shape the networks' predictions by analyzing how the probability distribution over class labels evolves with time, providing evidence that feedback connections are effective in diminishing the effects of occluders.

\section{Methods}
\subsection{Datasets}
To investigate the effects of occlusion in object recognition we present two novel stereoscopic image data sets.

\subsection*{Occluded Stereo Multi-MNIST}
\begin{figure}[hbtp]
\centering
\includegraphics[width=0.49\textwidth]{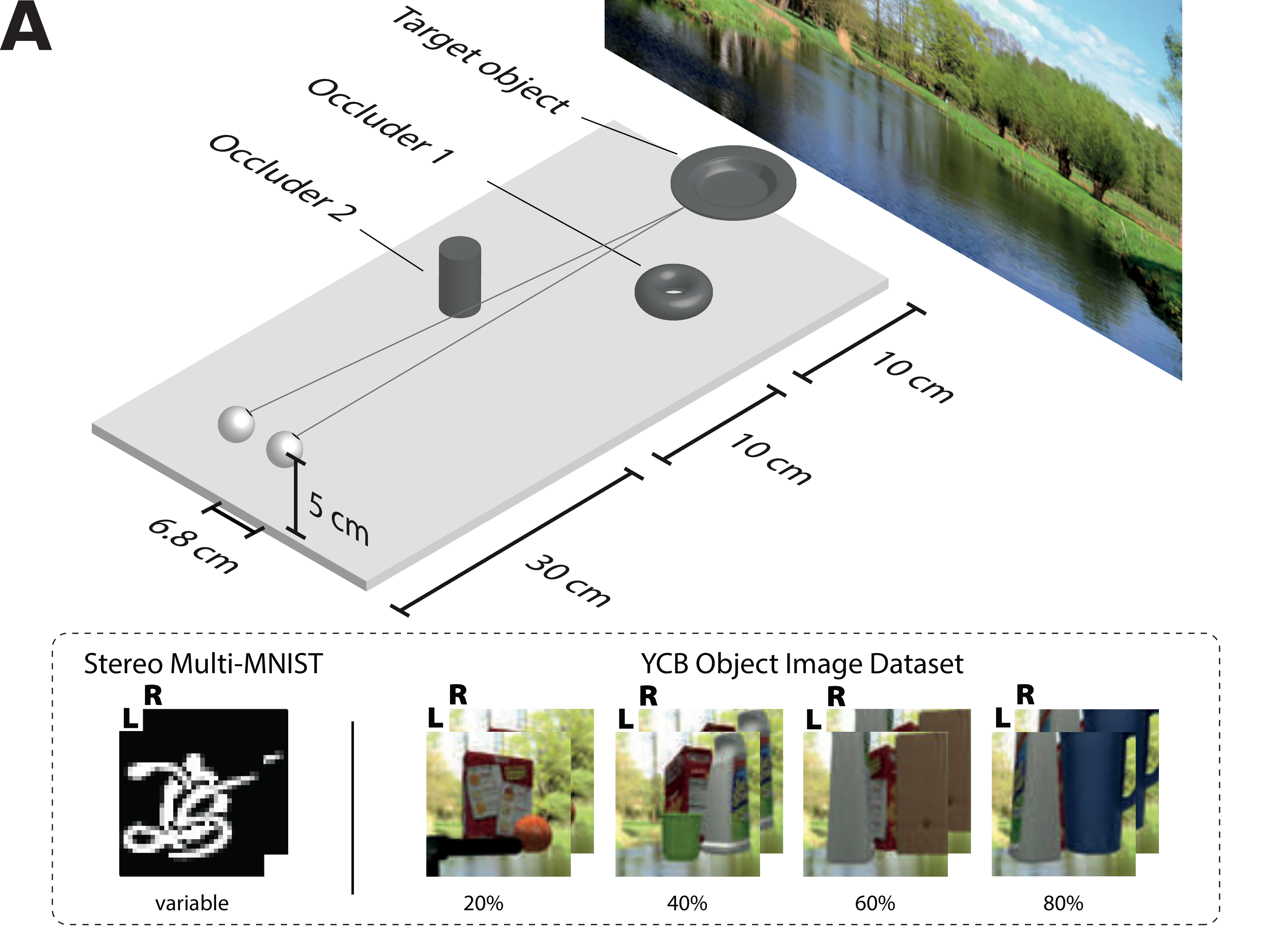}
\includegraphics[width=0.49\textwidth]{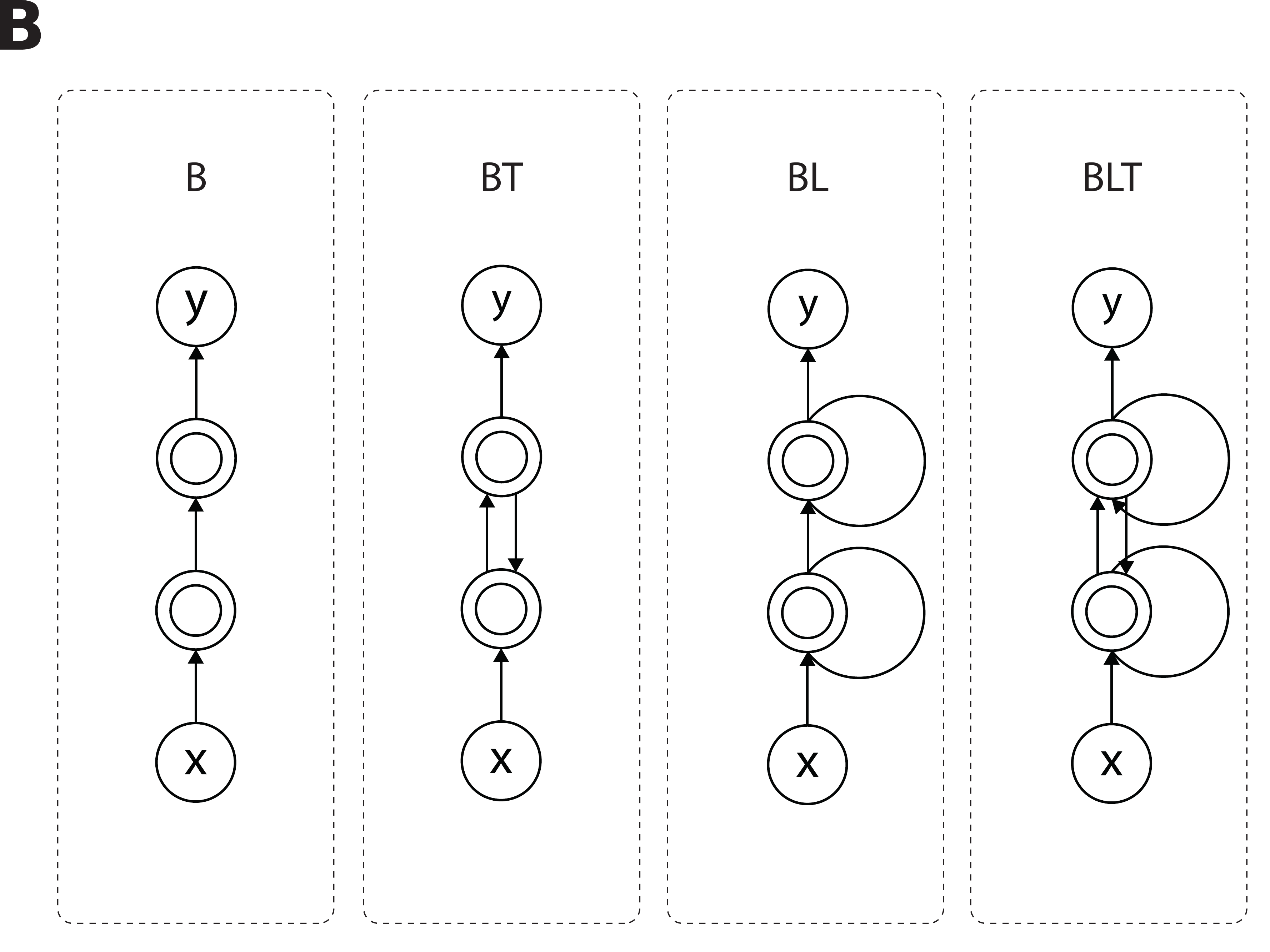}
\caption{The used stimuli and network models. (A) The centered target object is occluded by 2 occluder objects arranged in a 3D-fashion. (B) A sketch of the four network architectures named after their connection properties. B stands for bottom-up, L for lateral and T for top-down connections.}
\label{fig:network_overview}
\end{figure}

Occluded Stereo Multi-MNIST (OS-MNIST) is a stereoscopic occluded digit recognition dataset. We chose digit recognition because it is deemed a solved problem in computer vision, yet poses many of the challenges present in a realistic occlusion scenario. The use of MNIST digits additionally encourages the network to learn a representation that generalizes to different variants of a particular class \citep{lecun1998gradient}.
Contrary to past studies \citep{oreilly2013recurrent, smith2010nonstimulated, tang2014spatiotemporal, wyatte2012limits}, occlusion is generated by overlaying the target digit with other digit instances in a pseudo-3D environment. 
Every image of OS-MNIST contains three MNIST digit instances. Occlusion is generated by overlaying digits on top of each other as shown in Fig.~\ref{fig:network_overview} A.
The target object, i.e. the hindmost digit, is centered in the middle of the square canvas. Additional digits are then sequentially placed on top of the target object. These occluding objects remain fixed along the y-axis as if standing on a surface 5 cm below the viewer. The x-coordinate is drawn from a uniform distribution. The size of the digits was scaled to give the impression of objects with 20 cm height placed at different depths. We assumed a distance of 50 cm from the target object to the viewer, and 10 cm less for every added object. Images for the left and right eye were taken given an interocular distance of 6.8 cm. 
For each MNIST digit we generated 10 random occluder combinations of the remaining classes, resulting in a total of $600{,}000$ images for training and $100{,}000$ images for testing. All stimuli were rendered at $32 \times 32$ pixels. The generative source code for OS-MNIST will be made available. %\footnote{https://www.github.com/mrernst/os-mnist/}

\subsection*{Occluded Stereo YCB-Objects}
The Occluded Stereo YCB-Object image dataset (OS-YCB) constitutes the second novel dataset for stereoscopic occluded object recognition. OS-YCB contains stereo image stimuli of common household objects occluding each other. We chose 79 objects from an assortment of items suitable for robotics applications, i.e. the YCB-Object set \citep{calli2015benchmarking,calli2015ycb}. For each image, we placed three virtual 3D objects according to Fig.~\ref{fig:network_overview} and using a repurposed robot simulator as a stereoscopic camera \citep{metta2010icub}. The dataset is set to be publicly released alongside this publication.

In line with our other dataset, OS-MNIST, the objects are placed at a distance of 50~cm from the viewer on a plane 5~cm below line of sight, see Fig.~\ref{fig:network_overview} A. All objects are placed in upright position and turned by a random yaw angle, again encouraging the model to learn a representation that generalizes to different properties of a particular class. A background was chosen to simulate a context with natural image statistics. The occlusion percentage of each image is defined as the ratio of occluded to visible pixels averaged over the two stereo images.

We generated $1{,}000$ images per object and occlusion percentage (20, 40, 60, 80 \%) resulting in $316{,}000$ stereo images. The occluders were chosen in a way that no two instances of one class would appear in the same image. The images were rendered at $320 \times 240$, $160 \times 120$ and $80 \times 60$ pixels. For all experiments conducted we used a $32 \times 32$ pixel central crop of the $80 \times 60$ version. The data were randomly split into 80 \% training and 20 \% testing data.

\subsection{Network Models}

The three aforementioned connection types enable four basic network architectures as shown in Fig.~\ref{fig:network_overview} B: Bottom-up connection only (\emph{B}), bottom-up and top-down connections (\emph{BT}), bottom-up and lateral connections (\emph{BL}), and bottom-up, lateral, and top-down connections (\emph{BLT}). As lateral and top-down connections introduce cycles into the computational graph, these models represent recurrent neural networks and allow for information to be retained within a layer or to flow back into previous layers.

Each of the models consists of an input layer, two hidden recurrent layers and an output layer. Bottom-up and lateral connections are implemented as convolutional layers \citep{lecun1998gradient} with a stride of $1 \times 1$ followed by a $2 \times 2$ maxpooling operation with a stride of $2 \times 2$.
Top-down connections are implemented as transposed convolutions \citep{zeiler2010deconvolutional} with output stride $2 \times 2$ to match the input size of the convolutional layer that came before it. Each of the recurrent network models is unrolled and trained for four time steps by backpropagation \citep{rumelhart1986learning}. When reporting accuracy, the output at the last unrolled time step available for the particular architecture is used.
Recurrent network models naturally have more learnable parameters than their feedforward counterparts, due to their increased connectivity. To compensate for this, we introduce two additional feedforward models \emph{B-F} and \emph{B-K} as in \citep{spoerer2017recurrent}. \emph{B-F} doubles the number of convolutional filters in each of the hidden layers from 32 to 64. \emph{B-K} has larger $5 \times 5$ convolutional kernels compared to $3 \times 3$ of the standard \emph{B} model. The larger kernel size effectively increases the number of connections that each unit has and makes \emph{B-K} a more appropriate model for control. The additional feature maps in \emph{B-F} on the other hand alter the representational power of the model. The number of learnable parameters for each of the architectures can be found in Tab.~\ref{tab:networkparameters}.

\begin{table}[hbt]
%\small
\scriptsize

\centering
\caption{Number of learnable parameters for all models and input channels.}
\begin{tabular}{ccccccc}
\toprule
%\hline
%  &\multicolumn{6}{c}{Architecture} \\
%\cmidrule{2-7}
%\hline
 & \emph{B} & \emph{B-F}  & \emph{B-K} & \emph{BT} & \emph{BL} &  \emph{BLT} \\
\midrule
%\hline
Kernel size & $3 \times 3 $ &  $3 \times 3 $      &  $5 \times 5 $    &   $3 \times 3 $ &  $3 \times 3 $ &  $3 \times 3 $  \\
Hidden layer units  & $32$ & $64$      & $32$    &  $32$ & $32$ & $32$  \\
%Layer 2 units & $32$ & $64$      & $32$    &  $32$ & $32$ & $32$  \\
%Output units & $10$ & $10$      & $10$    &  $10$ & $10$ & $10$  \\
\midrule
%\hline
&\multicolumn{6}{c}{} \\
Image channels  &\multicolumn{6}{c}{Number of learnable parameters (OS-MNIST, 10 classes)} \\
\cmidrule{0-0} \cmidrule{2-7}
%\hline
1 & $9{,}898$ & $38{,}218$      & $26{,}794$     & $19{,}146$ &  $28{,}394$ & $37{,}642$  \\
2 & $10{,}186$ & $38{,}794$      & $27{,}594$     & $19{,}434$ &   $28{,}682$ & $37{,}930$  \\
&\multicolumn{6}{c}{} \\
Image channels  &\multicolumn{6}{c}{Number of learnable parameters (OS-YCB, 80 classes)} \\
\cmidrule{0-0} \cmidrule{2-7}

3 & $12{,}784$ & $43{,}920$ & $30{,}704$  & $22{,}032$ & $31{,}280$ & $40{,}528$ \\
6 & $13{,}648$ & $45{,}648$ & $33{,}104$  & $22{,}896$ & $32{,}144$ & $41{,}392$ \\
\bottomrule
%\hline
\end{tabular}

\label{tab:networkparameters}
\end{table}

\subsubsection{Layers}
The central element of the investigated models is the hidden recurrent convolutional layer. The inputs to these layer are denoted $\mathbf{h}_{i,j}^{(t,l)}$. This notation represents the vectorized input of a patch centered on location $(i,j)$ in layer $l$ computed at time step $t$ across all feature maps indexed by $k$. Thus an input stimulus presented to the network is denoted as $\mathbf{h}_{i,j}^{(t,0)}$. 
The activation $z$ of a hidden recurrent layer can then be written as
\begin{equation}
	z^{(t,l)}_{i,j,k} =\left( \mathbf{w}_{k}^{(l)B} \right)^\top \mathbf{h}^{(t,l-1)}_{i,j} + \left( \mathbf{w}_{k}^{(l)L} \right)^\top \mathbf{h}^{(t-1,l)}_{i,j} + \left( \mathbf{w}_{k}^{(l)T} \right)^\top \mathbf{h}^{(t-1,l+1)}_{i,j},% + b_{k}^{(l)}.
\end{equation}
where $\mathbf{w}_{k}^{(l)\cdot}$ is the vectorized convolutional kernel at feature map $k$ in layer $l$ for bottom-up (B), lateral (L), and top-down (T) connections, respectively. Each of these kernels only is active for architectures using the particular connection-type and is otherwise set to zero. Note that the lateral and top-down connections depend on values of the previous time step, thus we define the inputs to be a vector of zeroes for $t=0$, where there would be no preceding time step. Top-down connections are only present between the two hidden layers (Fig.~\ref{fig:network_overview} B).

Following the flow of information, the $z^{(t,l)}_{i,j,k}$ of the hidden layer are  batch-normalized \citep{ioffe2015batch}. This technique normalizes an activation $z$ using the mean $\mu_\mathcal{B}$ and standard deviation $\sigma_\mathcal{B}$ over a mini-batch of activations $\mathcal{B}$ and adds multiplicative and additive noise. % replacing the traditional bias.
\begin{equation}
	\mathrm{BN}_{\gamma, \beta}(z^{(t,l)}_{i,j,k})= \mathbf{\gamma}^{(l)}_k \cdot  \frac{z^{(t,l)}_{i,j,k} - \mathbf{\mu_\mathcal{B}}}{\mathbf{\sigma_\mathcal{B}}} + \mathbf{\beta}^{(l)}_k,
\end{equation}
where $\gamma$ and $\beta$ are additional learnable parameters. %for each connection type and layer.

The normalized activations then are passed on to rectified linear units (ReLU, $\sigma_z$) 
\begin{equation}
    \sigma_{z} \left(a^{(t,l)}_{i,j,k} \right) = \max \left( 0, a^{(t,l)}_{i,j,k} \right)
\end{equation}
and go through local response normalization (LRN, $\omega$)
\begin{equation}
	\omega(a^{(t,l)}_{i,j,k}) = a^{(t,l)}_{i,j,k} \left( c + \alpha \sum_{k' = \max(0, k-\frac{n}{2})}^{\min(n-1, k+\frac{n}{2})} \left(a^{(t,l)}_{i,j,k'}\right)^2 \right)^{-\beta},
\end{equation}
with $n=5$, $c=1$, $\alpha = 10^{-4}$ and $\beta = 0.5$. Inspired by lateral inhibition LRN induces competition for large activities amongst the $n$ closest features within a spatial location \citep{krizhevsky2012imagenet}.
Finally, the output $h^{(t,l)}_{i,j,k}$ of the hidden layer can be written as
\begin{equation}
	h^{(t,l)}_{i,j,k} = \omega \left( \sigma_z \left( \mathrm{BN}_{\mathbf{\gamma},\mathbf{\beta}} \left( z^{(t,l)}_{i,j,k}\right) \right) \right).
\end{equation}

After passing the second hidden layer the activations are relayed to a fully-connected segment with one output unit per class and a softmax activation layer, defined as:
\begin{equation}
	\mathrm{softmax}(\mathbf{a})_i = \frac{\exp(a_i)}{\sum_j \exp(a_j)}.
\end{equation}
This final activation function normalizes the sum over output units to one and therefore makes the output interpretable as the probability distribution over all possible classes.
%The resulting network output can be interpreted as the probability distribution over classes.

\subsubsection{Learning}
The labels to be predicted by the network are encoded as one-hot vectors. 
%Thus the target vector $\mathbf{y}$ is comprised of elements $y_i$ defined as
%\begin{equation}
%	 y_i=\begin{cases}
%               1 \text{ if } y_i =  \tilde{\mathrm{y}} \\ %\tilde{\vy} \tilde{\mathrm{y}} \tilde{\sY}
%               0 \text{ else}
%            \end{cases},
%\end{equation}
%where $\tilde{\mathrm{y}}$ the target object.
To quantify the mismatch between the networks' output $\hat{\mathbf{y}}^{(\tau)}$ and the the target label $\mathbf{y}$ we compute the cross-entropy cost-function summed across all $\tau$ time steps and all $N$ output units:
\begin{equation}
	J(\hat{\mathbf{y}}^{(\tau)}, \mathbf{y}) = - \sum_{t=0}^{\tau} \sum_{i=0}^{N} y_i \cdot \log \hat{y}^{(t)}_i + (1-y_i) \cdot \log(1- \hat{y}^{(t)}_i).
\end{equation}
The network parameters are adapted using adam \citep{kingma2014adam} with an initial learning rate of $\eta = 0.003$ to perform gradient descent. Unless stated otherwise training occurred for 25 epochs with mini-batches of size 400. Bottom-up weights were initialized with a truncated normal distribution with $\mu = 0$, $\sigma = \nicefrac{2}{\textrm{kernelsize}}$ and all other weights with $\mu = 0$, $\sigma = 0.1.$

\subsection{Model Performance Metrics}
The different models were evaluated in terms of classification accuracy averaged across the test set. We use pair-wise McNemar's tests \citep{mcnemar1947note} to compare test performances with each other. McNemar's test makes use of the variability in performance across stimuli for statistical inference and thus does not require repeated training \citep{dietterich1998approximate}. This method enables us to evaluate and compare a variety of different models in a computationally efficient manner.
As multiple comparisons increase the risk of false positives, we control the false discovery rate (FDR, the expected proportion of false positives among the positive outcomes) at 0.05 using a Bonferroni-type correction procedure developed in \citep{benjamini1995controlling}.

\section{Results}
\subsection{Model Performance}
\begin{figure}[hbtp]
\centering
\includegraphics[width=0.49\textwidth]{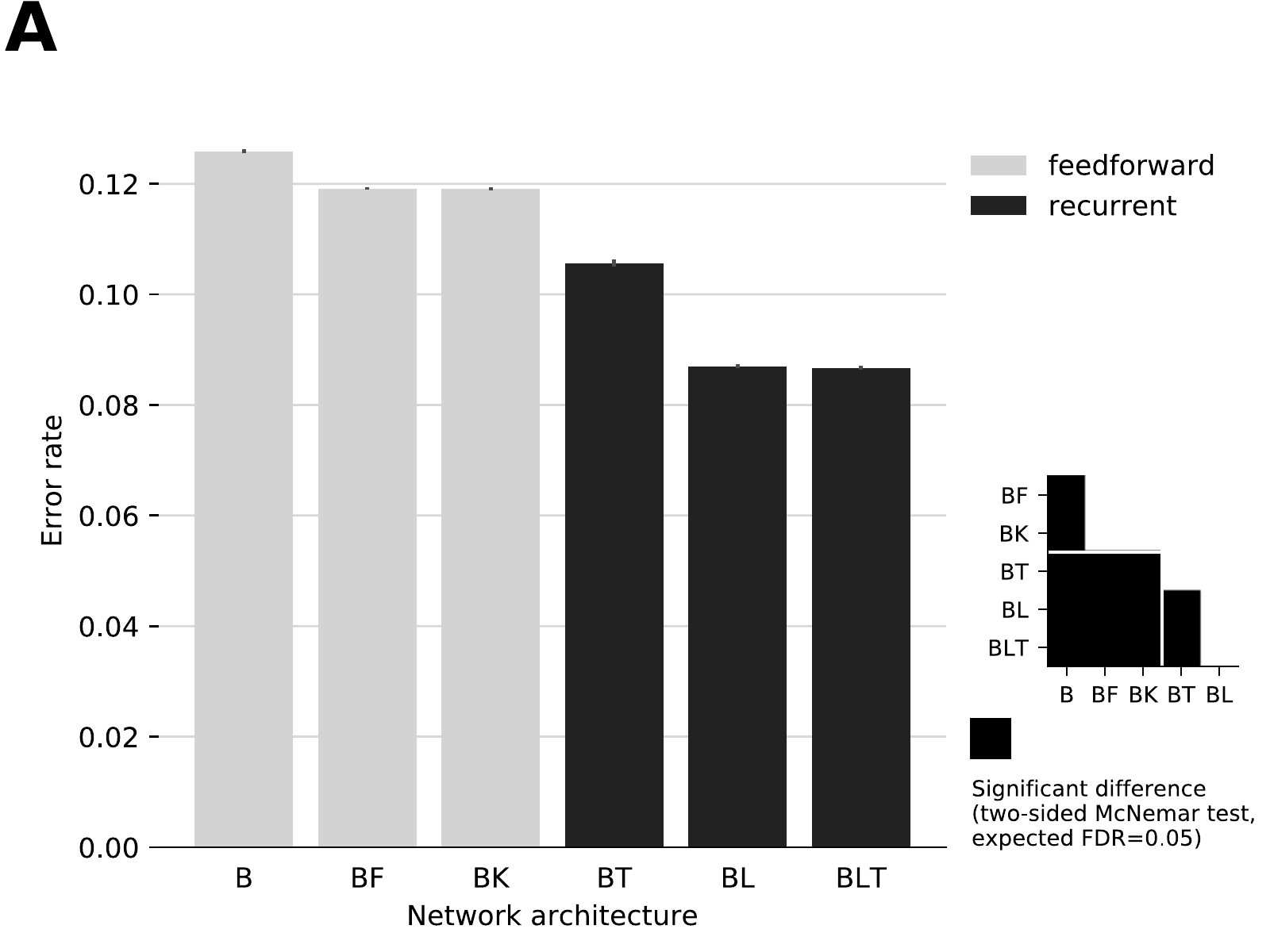}
\includegraphics[width=0.49\textwidth]{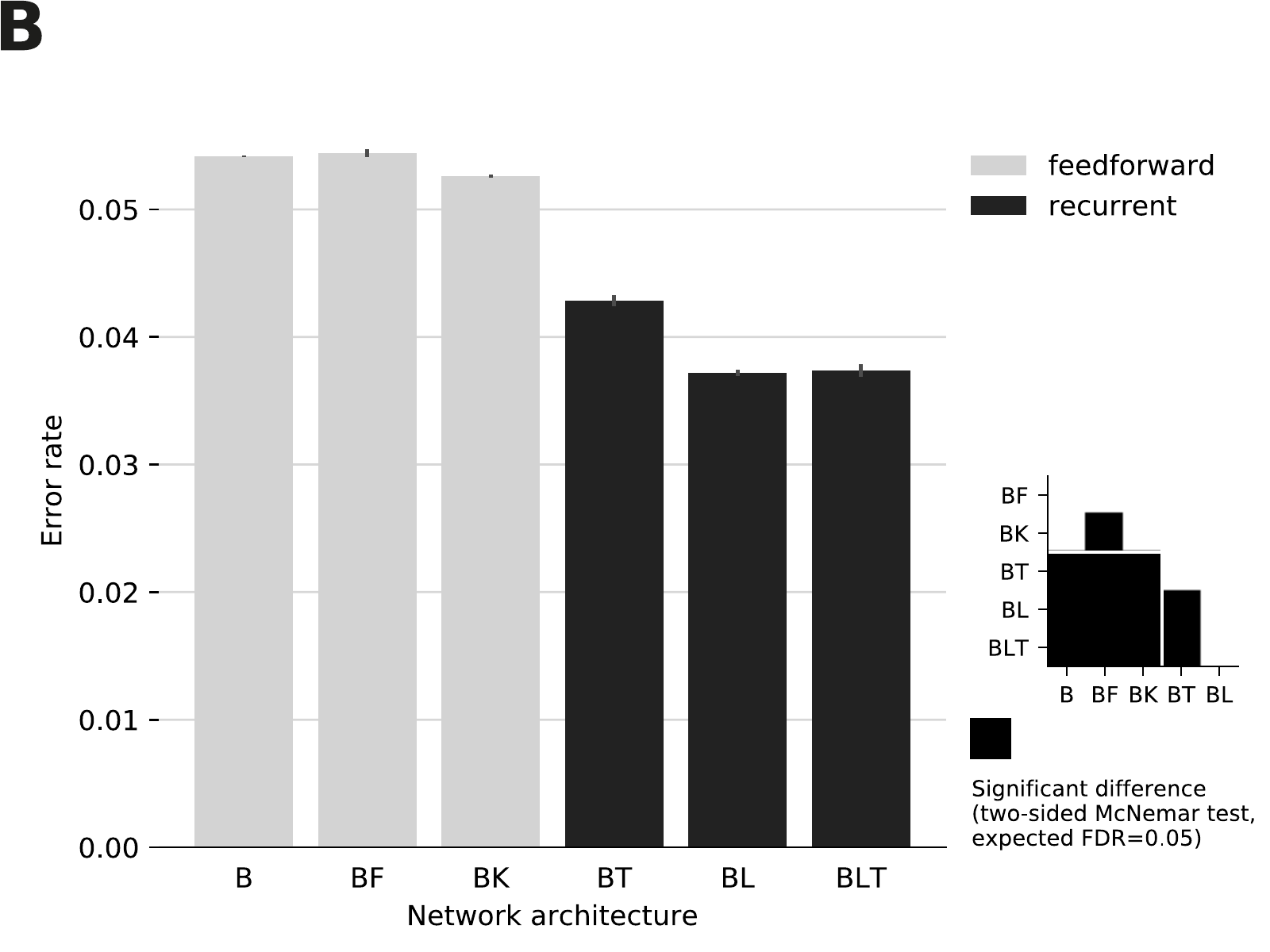}
\includegraphics[width=0.49\textwidth]{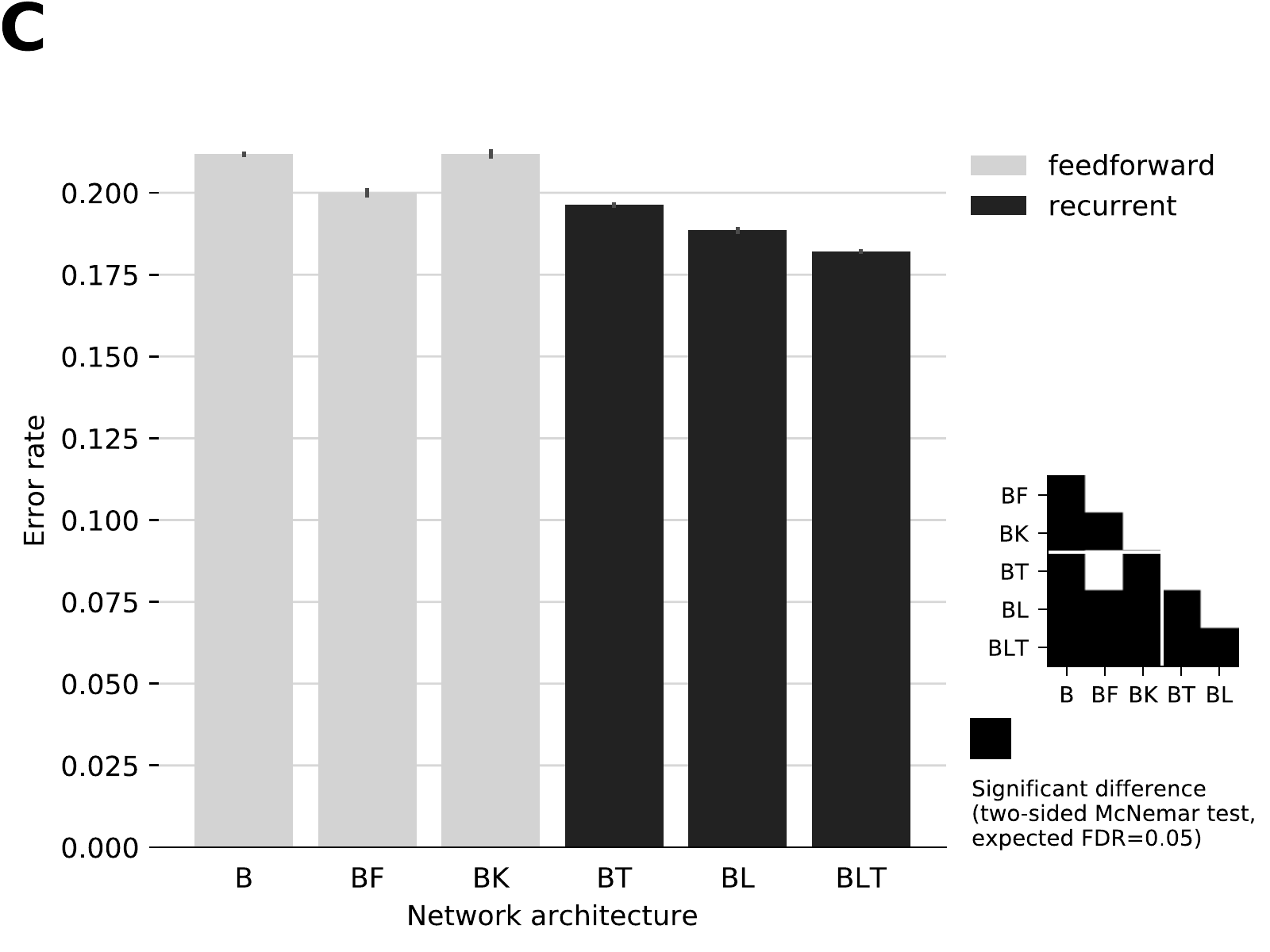}
\includegraphics[width=0.49\textwidth]{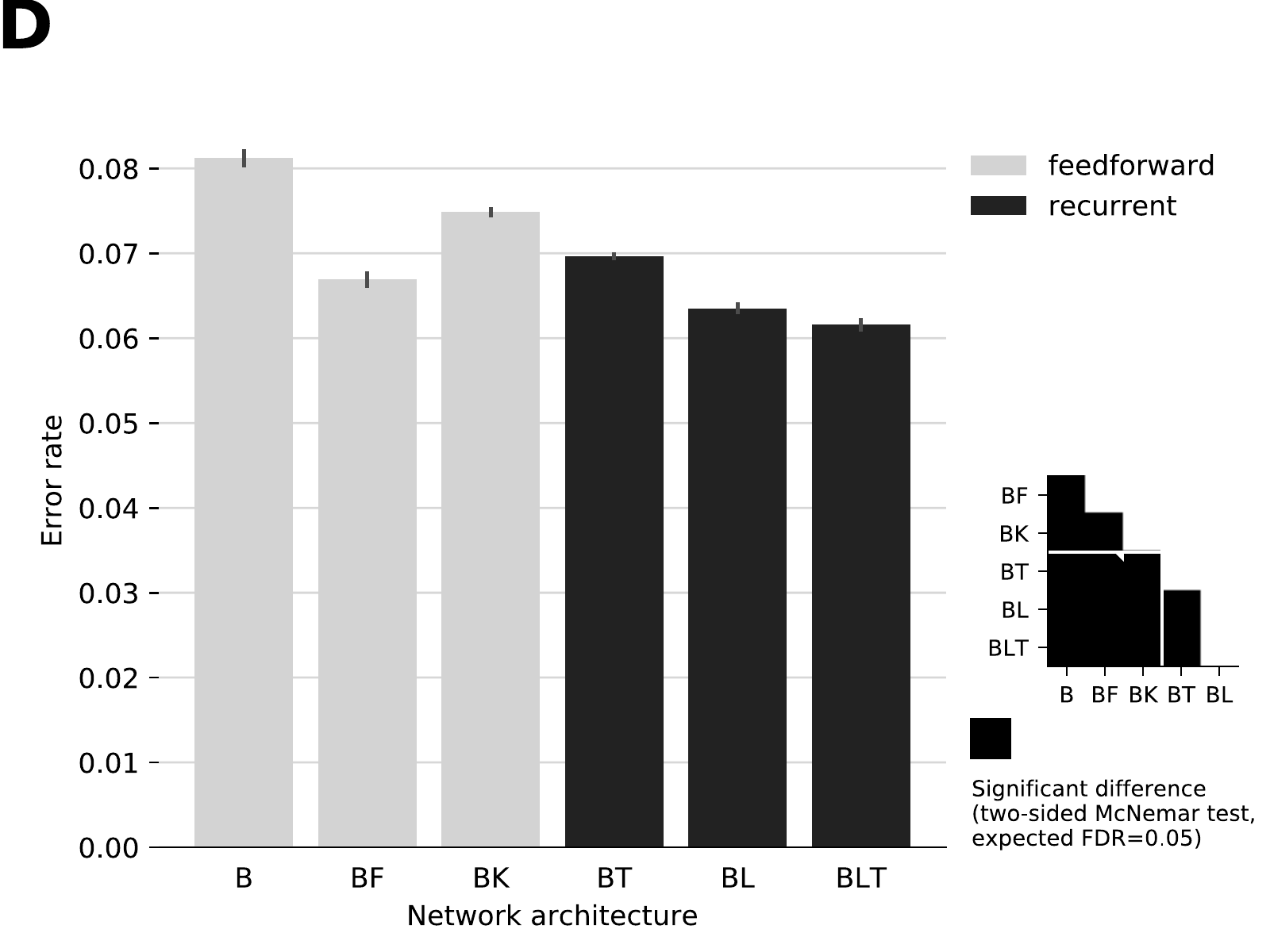}

\caption{Performance comparison of different network architectures. Error bars indicate the standard error based on five repetitions of the training and testing procedure. Matrices depict results of pairwise McNemar tests,  black squares indicating significant differences at $p < 0.05$. (A) OS-MNIST, monocular input. (B) OS-MNIST, binocular input. (C) OS-YCB, monocular input. (D) OS-YCB, binocular input.}
\label{fig:performance_results}
\end{figure}

Every considered model was trained for 25 epochs on OS-MNIST and on OS-YCB with all four occlusion percentage levels combined. Training was conducted using a NVIDIA Tesla K40c GPU. % (?additional details autumnchat node)
Fig.~\ref{fig:performance_results} depicts the error-rate $(1 - \textrm{accuracy})$ for the models trained with monocular (A, C) and stereoscopic (B, D) input.

While the task to be accomplished by the network is the same for both datasets, OS-YCB offers 79 possible classes compared to only 10 in OS-MNIST. Additionally, OS-YCB requires the models to recognize 3D objects shown from different angles, making it the more complex of the two datasets. Given that we didn't change the architectural complexity, better overall performance was to be expected for OS-MNIST. The error rates for OS-MNIST (mono., range: .087 -- .126) and OS-YCB (mono., range: .182 -- .212) confirm this assumption.  
%TODO: Make sure you compare the two datasets. Which one is the harder one? OS-MNIST or OS-YCB? Was this to be expected?

Overall, recurrent architectures show increased performance on the given task compared to feedforward networks of near-equal complexity.
Significant differences (FDR = 0.05) for OS-MNIST with monocular inputs can be attested for all combinations except (\emph{B-K}, \emph{B-F}), $\chi^2(1,N = 100{,}000) = 0.17, p = .68$ and (\emph{BL}, \emph{BLT}), $\chi^2(1,N = 100{,}000) = 0.12, p = .73$. 

The lower left $3 \times 3$ square, highlighted by a white line, indicates that all but two pair-wise tests between feedforward and recurrent models show a significant performance gain for recurrent architectures. 
Only for OS-YCB data  (Fig.~\ref{fig:performance_results} C, D) \emph{BT} does not significantly outperform \emph{B-F}, $\chi^2(1,N = 63{,}200) = 0.68, p = .41$. For stereoscopic input \emph{B-F} outperforms \emph{BT}, however \emph{BLT} still performs best. Notably, \emph{B-K} performs significantly worse than \emph{B-F} questioning the benefits of the increased kernel size, $\chi^2(1,N = 63{,}200) = 8.58, p = .02$.

Qualitatively, we observe similarities when comparing OS-MNIST and OS-YCB performances: Recurrent networks reliably outperform their non-recurrent counterparts and \emph{BT} produces the highest error-rates amongst recurrent models. Notably, the relative decrease in error-rate from feedforward to recurrent models is elevated for the stereoscopic case in OS-MNIST.
However, significant differences between \emph{BL} and \emph{BLT} can only be observed for the more complex OS-YCB data. Also, the relative performance of \emph{B-F} is substantially better for the OS-YCB dataset.
When trained separately on the four subsets of OS-YCB with specific occlusion percentages, we observe the same overall patterns. Error rates rise with higher occlusion percentages, but the BLT model consistently ranks highest in classification performance, see Tab.~\ref{tab:performance_comparison_sdd}.

\begin{table}[hbt]
\scriptsize
\centering
\caption{Error rates for all model architectures, standard error based on five independent training runs. 20, 40, 60, 80 \% occ. runs were trained for 100 ep., batchsize 100. Best performance per dataset is highlighted in bold.}
\begin{tabular}{cccccccc}

 &\multicolumn{7}{c}{OS-MNIST} \\

\toprule
%\cmidrule{3-8}
 & \% occ. & \emph{B} & \emph{B-F}  & \emph{B-K} & \emph{BT} & \emph{BL} &  \emph{BLT} \\
\midrule

\multirow{1}{*}{Mono}
&all& $.126 \pm .000$ & $.119 \pm .000$      & $.119 \pm .000$    &  $.106 \pm .001$ & $\mathbf{.087 \pm .000}$ & $\mathbf{.087 \pm .000}$  \\
\cmidrule{1-8}
\multirow{1}{*}{Stereo}
&all& $.054 \pm .000$ & $.054 \pm .000$      & $.053 \pm .000$    &  $.043 \pm .000$ & $\mathbf{.037 \pm .000}$ & $\mathbf{.037 \pm .000}$  \\
\bottomrule
\\

 &\multicolumn{7}{c}{OS-YCB} \\
\toprule
%\cmidrule{3-8}
 & \% occ. & \emph{B} & \emph{B-F}  & \emph{B-K} & \emph{BT} & \emph{BL} &  \emph{BLT} \\
\midrule

\multirow{5}{*}{Mono}
&20& $.039 \pm .001$ & $.036 \pm .001$      & $.039 \pm .001$    &  $.037 \pm .001$ & $.034 \pm .001$ & $\mathbf{.033 \pm .001}$  \\
&40& $.080 \pm .001$ & $.069 \pm .001$      & $.077 \pm .001$    &  $.075 \pm .001$ & $.069 \pm .001$ & $\mathbf{.067 \pm .001}$  \\
&60& $.194 \pm .001$ & $.173 \pm .001$      & $.193 \pm .001$    &  $.179 \pm .001$ & $.169 \pm .001$ & $\mathbf{.163 \pm .001}$  \\
&80& $.537 \pm .002$ & $.517 \pm .001$      & $.528 \pm .002$    &  $.510 \pm .002$ & $.508 \pm .002$ & $\mathbf{.502 \pm .002}$  \\
&all& $.212 \pm .002$ & $.200 \pm .001$      & $.212 \pm .001$    &  $.196 \pm .001$ & $.189 \pm .001$ & $\mathbf{.182 \pm .001}$  \\
\cmidrule{1-8}
\multirow{5}{*}{Stereo}
&20& $.026 \pm .001$ & $.024 \pm .001$      & $.027 \pm .000$    &  $.025 \pm .001$ & $\mathbf{.022 \pm .000}$ & $\mathbf{.022 \pm .001}$  \\
&40& $.044 \pm .000$ & $.039 \pm .001$      & $.044 \pm .001$    &  $.042 \pm .000$ & $.039 \pm .001$ & $\mathbf{.037 \pm .000}$  \\
&60& $.083 \pm .001$ & $.070 \pm .001$      & $.081 \pm .002$    &  $.077 \pm .001$ & $.070 \pm .001$ & $\mathbf{.069 \pm .001}$  \\
&80& $.182 \pm .003$ & $.161 \pm .003$      & $.172 \pm .002$    &  $.167 \pm .001$ & $.154 \pm .002$ & $\mathbf{.153 \pm .003}$  \\
&all& $.081 \pm .001$ & $.067 \pm .001$      & $.075 \pm .001$    &  $.070 \pm .000$ & $.064 \pm .001$ & $\mathbf{.062 \pm .001}$  \\
\bottomrule
\end{tabular}

\label{tab:performance_comparison_sdd}
\end{table}

%\begin{table}[hbt]
%\scriptsize
%\centering
%\caption{Accuracy for all model architectures, standard error based on five independent training runs. 2, 3, 4 occ. runs were trained for 100 ep., batchsize 100. Best performance per dataset is highlighted in bold.}
%\begin{tabular}{cccccccc}
% &\multicolumn{7}{c}{Stereo Multi MNIST} \\
%
%\toprule
%%\cmidrule{3-8}
% & \% occ. & \emph{B} & \emph{B-F}  & \emph{B-K} & \emph{BL} & \emph{BT} &  \emph{BLT} \\
%\midrule
%
%
%\multirow{1}{*}{mono}
%&all& $.000 \pm .000$ & $.000 \pm .000$      & $.000 \pm .000$    &  $.000 \pm .000$ & $.000 \pm .000$ & $\mathbf{.000 \pm .000}$  \\
%\cmidrule{1-8}
%\multirow{1}{*}{stereo}
%&all& $.000 \pm .000$ & $.000 \pm .000$      & $.000 \pm .000$    &  $.000 \pm .000$ & $.000 \pm .000$ & $\mathbf{.000 \pm .000}$  \\
%\bottomrule
%\end{tabular}
%
%\label{tab:performance_comparison_sdd}
%\end{table}

\subsection{Impact of Recurrent Connectivity}
\begin{figure}[hbtp]
\centering
\includegraphics[width=1.00\textwidth]{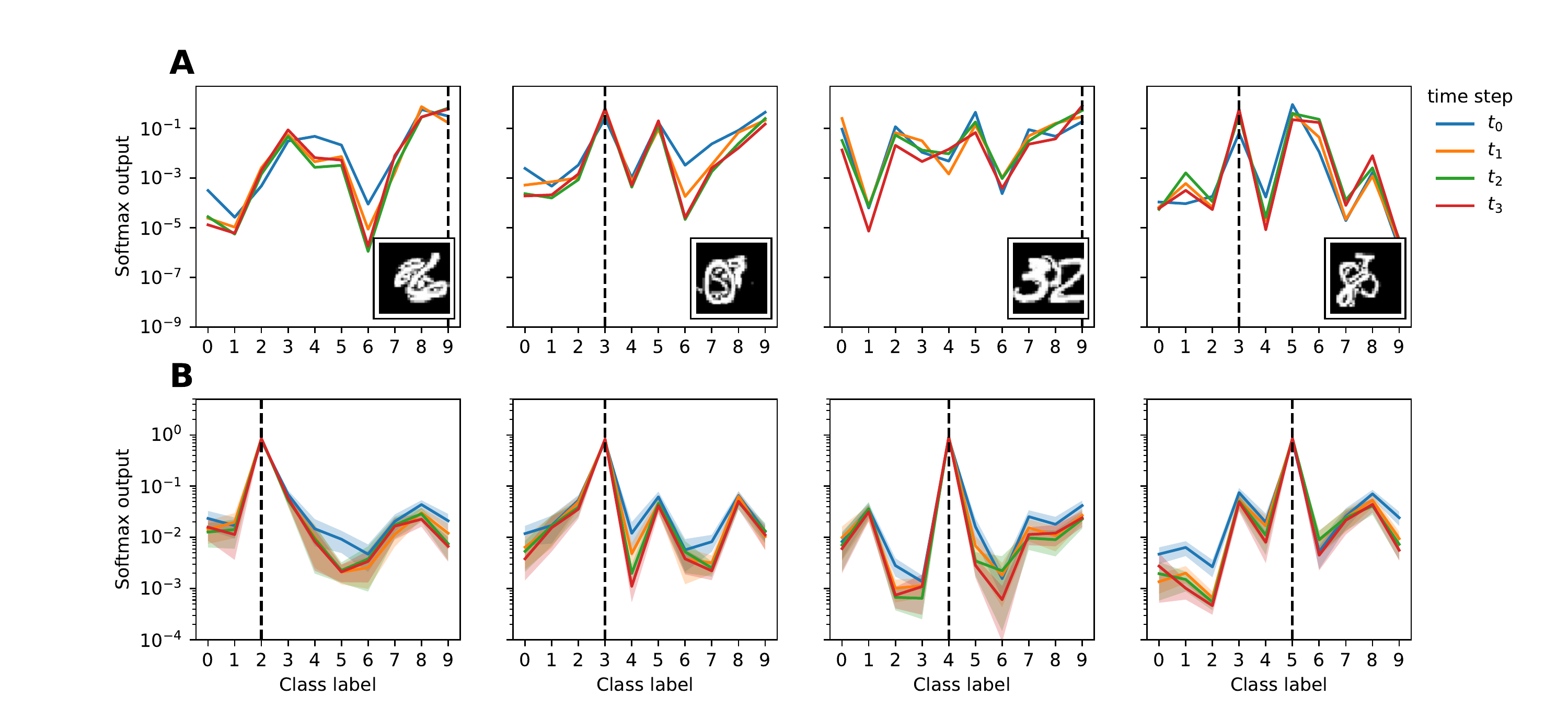}
\caption{Softmax output of \emph{BLT}. (A) Specific stimuli illustrating the effect of recurrent feedback. (B) Mean softmax output over all test stimuli of specific classes (2, 3, 4, 5). Shaded areas correspond to standard error estimated with a sample size of $1{,}000$ images taken from the test set.}
\label{fig:recurrent_visualizationA}
\end{figure}

The softmax output indicates how confident the network ``feels'' about each class being the target. Thus, evaluating the output of \emph{BLT} at each time step yields insight into how recurrent connections revise the networks' belief over time. In fact, we observe that wrong initial guesses at $t_0$ can be corrected at later time steps and correct initial guesses are reinforced. Examples of this can be found in Fig.~\ref{fig:recurrent_visualizationA} A: While the network initially estimates the target to be 8, the final output is the correct class 9 (left panel). The average softmax output over each target is depicted in Fig.~\ref{fig:recurrent_visualizationA} B. The visualization reveals that the probabilities assigned to incorrect classes decrease over time and also shows that the model has discovered systematic similarities between digits 3, 5 and 8, and digits 4 and 1.

% ME: TSNE macht nur bedingt Sinn, weil für die Digits keinen kanonischen un-occluded Stimulus haben. Für die Zukunft können wir vielleicht einen gemittelten Input über die Klassen errechnen und dann mit ins Netzwerk geben, aber ich fürchte bis morgen wird mir das nicht gelingen.

%To better understand how recurrent connections revise the internal representation of the network and therefore contribute to the performance gains we back-propagated the mismatch between hidden activations at different times to the input while keeping all other weights constant. Thus we tried to find a new input $x'$ that would cause the same hidden activation at time step $0$ as the original input $x$ did at time step $\tau$, or formally:
%\begin{equation}
%	a^{(0)}(x') \overset{!}{=} a^{(\tau)}(x). 
%\end{equation}
%We used the mean-squared error and adam.
%Fig.~\ref{fig:recurrent_visualizationB} illustrates the reconstructed inputs resulting from the error minimization. One can see...? 
%
%
%\begin{figure}[hbtp]
%\centering
%\includegraphics[width=.49\textwidth]{./figures/hidden_mismatch.pdf}
%\includegraphics[width=.49\textwidth]{./figures/reconstruction_hold.pdf}
%\caption{Recurrent Visualization B (Saliency Maps or Backpropagated Results)}
%\label{fig:recurrent_visualizationB}
%\end{figure}

\section{Discussion}

We investigated whether recurrent connectivity benefits occluded object recognition. Previous attempts at answering this question have been limited by using very simplistic and unnatural stimuli. On the one hand, the stimuli used by \citep{spoerer2017recurrent} were computer rendered digits without any variability in individual digit appearance. On the other hand, the stimuli used by \citep{oreilly2013recurrent, tang2014spatiotemporal} only blurred out image parts rather than introducing occluding objects. To overcome these limitations, we introduced two novel datasets that capture the natural variability of object appearance and a range of disparity and perspective cues, namely the Occluded Stereo-Multi MNIST (OS-MNIST) and the Occluded Stereo YCB-Objects (OS-YCB) datasets. We demonstrated that feedback connections significantly improve occluded object recognition for these much more complex datasets, providing strong evidence for a general benefit of recurrence for occluded object recognition. Furthermore, recurrent architectures similar to the ones presented here have been shown to also outperform parameter matched control models when no occlusion is present \citep{liang2015recurrent, han2018deeppredictive}, suggesting a rather general benefit of recurrence for object recognition tasks. This would be consistent with biological observations of how object information in the brain unfolds over time during recognition \citep{brincat2006dynamic}. %ME: ja, ich finde das kann man so schreiben.

%Contrary to the findings of \cite{spoerer2017recurrent} \emph{B-K} did not reliably outperform the \emph{B-F} model. The reported advantage of larger receptive fields might be linked to stimuli where certain irregularities of the occluders only become obvious at larger scales.

% Of the feedforward ensemble \emph{B-F} performed best on the more complex OS-YCB data. We assume that the additional convolutional filters help in situations where an object needs to be recognized from different angles, thus making use of the additional representational power. 

%The near-equal performance of \emph{BL} and \emph{BLT} questions the overall benefit of top-down connections. One of the challenges the single top-down layer has to overcome is being the servant of two masters. One the one hand it is responsible of moving the representation closer to its final destination, on the other hand it is supposed to transfer higher order . -> I don't remember anymore, ask Thomas.

In our experiments, the fully recurrent model comprising lateral and top-down connections (\emph{BLT}) performed best in all runs. The \emph{BL} model came in second, while the \emph{BT} performed worst, suggesting that lateral connections are particularly important for the observed performance advantage.

A second finding is that for stereoscopic input we observed higher recognition rates. This is likely due to the fact that the additional image introduces a new perspective of the scene, potentially revealing additional information about the target. Furthermore, for the stereoscopic input, the target is presented at zero disparity, while the occluder objects are not, which provides the network with a cue regarding what part of the input should be ignored.
Qualitatively, the results of the statistical network comparisons resemble the ones obtained for monocular stimuli. Interestingly, however, the relative performance difference between recurrent and feedforward models was usually higher for stereoscopic stimuli. This suggests that the recurrent connections are effective in utilizing the additional cues provided by the binocular presentation of the scene. %leider alles nicht so schön wie beim ICANN paper, wo auch die absolut Differenzen größer für Stereoskopischen Input waren. Relativ gesehen stimmt das aber immer noch.

During training, we consistently observed that the sum of recurrent weights (lateral and top-down) became slightly negative. We hypothesize that this bias towards negative weights might also contribute to inhibiting or discounting occluders. With the network's  dynamics being determined by the ReLU activation function a slight bias towards inhibitory weights might also be important for keeping activations centered around the non-linearity. Finally, our analysis revealed that recurrent connections revise the network's output over time, sometimes correcting an incorrect initial output after the first feedforward pass through the network, providing further evidence for the effectiveness of recurrence.

Evidently, any recurrent computation could also be performed by an appropriately unfolded (and therefore deeper) feedforward network. The recurrent network can be viewed as equivalent to such a deeper feedforward version, with certain weights constrained to be identical. Thus, recurrence implies a form of weight sharing in the temporal domain similar to how convolutional layers implement a form of weight sharing in the spatial domain. We speculate that this is the chief reason for the observed performance gains of recurrent networks \citep{liao2016residual}. % JT: has anybody else claimed this before? Would be bad not to cite them here!

In conclusion, we have shown that recurrent neural network architectures show significant advantages on complex occluded object recognition problems. Given their improved performance and greater biological plausibility they deserve more thorough analysis.

%In conclusion, recurrent convolutional neural networks have been shown to outperform feedforward networks at occluded object recognition.
%As previous works were severely limited by using very simplistic stimuli \citep{spoerer2017recurrent, oreilly2013recurrent}, we introduced two novel datasets that capture the full range of disparity and perspective cues, namely Stereo-Multi MNIST and YCB object image dataset. For the first time, we could demonstrate that recurrent connectivity benefits occluded object recognition for complex binocular stimuli, which approach natural object recognition. In fact, the advantages were even greater for the more realistic stereoscopic input compared to monocular input. 
%Building on previous work, our results pave the way for potential applications and provide further evidence for the necessity of recurrent neural networks in the domain of occluded object recognition.

\subsubsection*{Acknowledgments}

This work was supported by the European Union’s Horizon 2020 research and innovation programme under grant agreement No. 713010  (GOAL-Robots, Goal-based Open-ended Autonomous Learning Robots).
%\textbf{Not to be included in the anonymous submission!!}

\bibliographystyle{abbrv}
\bibliography{nips.bib}
\end{document}